\documentclass[11pt]{article}

\usepackage[final]{acl}

\usepackage{times}
\usepackage{latexsym}

\usepackage[T1]{fontenc}

\usepackage[utf8]{inputenc}

\usepackage{microtype}

\usepackage{inconsolata}

\usepackage{graphicx}

\usepackage{booktabs}
\usepackage{tabularx}
\usepackage{multirow}
\usepackage{array}
\usepackage{algorithm}
\usepackage{algpseudocode}
\usepackage{amsmath}
\usepackage{amssymb}

\usepackage{stfloats}
\usepackage{pgfplots}
\pgfplotsset{compat=1.18}

\usepackage{hyperref}
\usepackage[most,skins]{tcolorbox} 
\usepackage{enumitem}
\usepackage[table]{xcolor}
\tcbuselibrary{skins,breakable}
\usepackage{tabularx}
\usepackage{array}
\usepackage{algorithm}
\usepackage{algpseudocode}
\definecolor{citebluish}{HTML}{397fbe}
\hypersetup{
  colorlinks=true,
  linkcolor=citebluish,
  citecolor=citebluish,
  urlcolor=citebluish
}

\definecolor{mygreen}{HTML}{298D66}
\definecolor{myred}{HTML}{F94153}
\definecolor{tts_or}{HTML}{8AD879}
\definecolor{tts_sc}{HTML}{53D2DC}
\definecolor{grpo}{HTML}{3C5488}
\definecolor{gspo}{HTML}{8491B4}
\definecolor{srpo}{HTML}{4DBBD5}
\definecolor{octopus}{HTML}{669999}
\newcolumntype{L}{>{\hspace{2.2pt}}l<{\hspace{2.2pt}}}
\newcolumntype{Y}{>{\hspace{2.2pt}\centering\arraybackslash}X<{\hspace{2.2pt}}}

\tcbset{
  promptbox/.style={
    width=\linewidth,
    top=8pt,
    bottom=2pt,
    left=4pt,
    right=4pt,
    colback=gray!1!white,
    colframe=black,
    colbacktitle=black,
    enhanced,
    center,
    fontupper=\normalsize,
    before upper={\setlength{\baselineskip}{1.12\baselineskip}}, 
    attach boxed title to top left={yshift=-0.1in,xshift=0.15in},
    boxed title style={boxrule=0pt,colframe=white,},
  }
}

\newtcolorbox{promptbox}[2][]{promptbox,title=#2,#1}

\usepackage{tcolorbox} 
\tcbuselibrary{breakable,skins,listings}

\usepackage{listings}   
\definecolor{verifierpurple}{RGB}{226,214,241}
\definecolor{verifierdarkpurple}{RGB}{171,94,232}
\colorlet{verifierpurple}{black!4}   
\colorlet{verifierdarkpurple}{black} 
\tcbset{
  boxrule=0.5pt,
  left=2pt,right=2pt,top=2pt,bottom=2pt,
  before skip=6pt, after skip=6pt
}
\newtcolorbox{verifierbox}[2][]{%
  enhanced, breakable,
  colback=verifierpurple!25,
  colframe=verifierdarkpurple,
  colbacktitle=verifierdarkpurple,
  coltitle=white,
  fonttitle=\bfseries,
  title={#2},
  #1
}

\title{\textsc{SkillPyramid}: A Hierarchical Skill Consolidation Framework for Self-Evolving Agents}

\author{
\textbf{Yuan Xiong}\textsuperscript{1,2}\thanks{Equal contribution.} \quad
\textbf{Ziqi Miao}\textsuperscript{3}\footnotemark[1] \quad
\textbf{Chen Qian}\textsuperscript{3} \quad
\textbf{Lijun Li}\textsuperscript{3} \\
\textbf{Ken Deng}\textsuperscript{5} \quad
\textbf{Yequan Wang}\textsuperscript{4}\thanks{Corresponding author.} \quad
\textbf{Shizhu He}\textsuperscript{1,2,4}\footnotemark[2] \\
\textbf{Jun Zhao}\textsuperscript{1,2} \quad
\textbf{Kang Liu}\textsuperscript{1,2}
\\[0.6em]
\textsuperscript{1}School of Artificial Intelligence, University of Chinese Academy of Sciences, Beijing, China \\
\textsuperscript{2}The Key Laboratory of Cognition and Decision Intelligence for Complex Systems, \\
Institute of Automation, Chinese Academy of Sciences, Beijing, China \\
\textsuperscript{3}Shanghai Artificial Intelligence Laboratory, Shanghai, China \\
\textsuperscript{4}Beijing Academy of Artificial Intelligence, Beijing, China \\
\textsuperscript{5}Independent Researcher
\\[0.6em]
\texttt{\{shizhu.he,jzhao,kliu\}@nlpr.ia.ac.cn}
}

\begin{document}
\maketitle

\begin{abstract}
Recent AI agents can flexibly invoke skills to solve complex tasks, but their long-term improvement is fundamentally constrained by a lack of systematic skill construction, accumulation, and transfer. In particular, without a unified framework for skill consolidation, agents tend to redundantly construct similar capabilities across different tasks, are unable to effectively transform experience into reusable assets, and struggle to generalize task-specific skills to novel scenarios. To address this limitation, we propose \textsc{SkillPyramid}, a skill consolidation framework that reuses existing skill experience for broader task generalization. Operating on a hierarchical skill topology, \textsc{SkillPyramid} further introduces a self-evolution mechanism that enables agents to compose, validate, and incorporate new skills during task execution. Experiments on ALFWorld, WebShop, and ScienceWorld across four backbone models show that \textsc{SkillPyramid} substantially increases the average reward by 38.0\% and reduces execution steps by 27.7\%. Overall, our method transforms a skill collection from a static resource pool into a dynamic evolution system.
Code is available at \url{https://github.com/xiongyuaay/skill_pyramid}.
\end{abstract}

\section{Introduction}
\begin{figure}[t]
    \centering
    \includegraphics[width=1\linewidth]{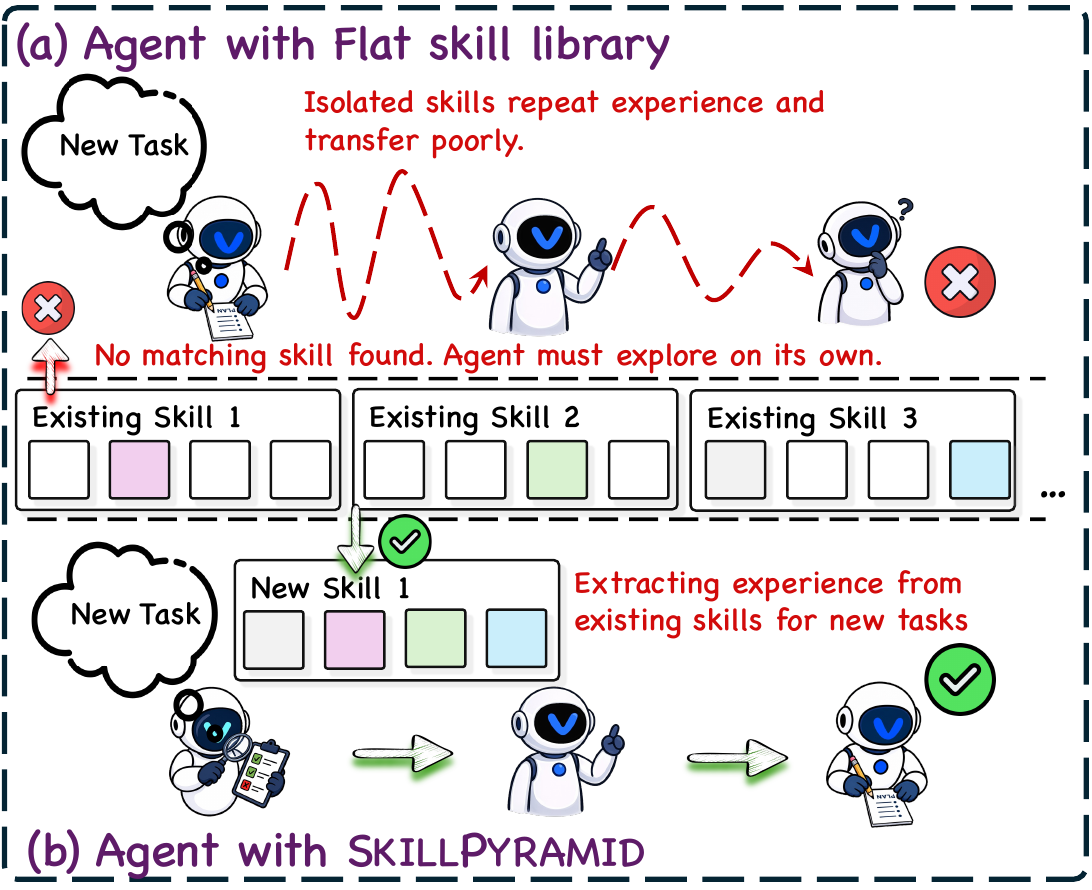}
    \caption{Flat skill library vs.\ \textsc{SkillPyramid}.
(a) With isolated skills, no match is found for an unseen task and the agent must explore from scratch, often failing.
(b) \textsc{SkillPyramid} composes new skills by recombining components of existing ones.}
    \label{fig:intro}
\end{figure}

Recent advances in large language models (LLMs) have shifted agent systems from passive response generation to autonomous task completion~\citep{yao2022react,shinn2023reflexion,wang2023voyager}. Within this shift, \emph{skills}—callable units of capability—have become a key substrate for agents' problem solving, substantially improving performance on complex, long-horizon, and open-ended tasks~\citep{wang2023voyager,liang2026skillnet,wang2026skillx,mi2026skillpro}.
In open-ended environments, however, manually curated skill sets quickly grow stale, and continual skill evolution becomes essential for sustained competence.
Yet a central question remains open: how can skills evolve from isolated, task-specific solutions into reusable, compositional knowledge?

Existing approaches tackle parts of this question from different angles.
Tool-augmented agents extend LLMs with external APIs to fill capability gaps~\citep{schick2023toolformer,qin2023toolllm}, and modular frameworks coordinate specialized components through task decomposition~\citep{karpas2022mrkl,shen2023hugginggpt,lu2023chameleon}.
A related line learns reusable skills via program synthesis or experiential accumulation~\citep{liang2022codepolicies,zhao2023expel,fang2025memp}.
More recently, dedicated skill-management frameworks have emerged~\citep{liang2026skillnet,wang2026skillx,mi2026skillpro}, demonstrating the value of creating, organizing, retrieving, and refining reusable skill units.
Yet two limitations remain: (i) existing skills are typically reused as stored units, leaving their internal transferable experience underexploited; and (ii) new skills are often retrieved, refined, or appended rather than composed from prior skill experience.

This motivates a different perspective: a growing, structured skill library in which the experience captured by existing skills generalizes effectively to tasks beyond their original scope. As illustrated in Figure~\ref{fig:intro}, related skills often share underlying capabilities, and treating these as atomic components lets agents recombine them to address unseen tasks, turning a flat collection into a structure that supports both reuse and generalization.

Building on this view, we propose \textsc{SkillPyramid}, an evolving skill framework for LLM agents. \textsc{SkillPyramid} organizes skills into a multi-level hierarchy in which lower levels hold fine-grained, reusable atomic capabilities and upper levels abstract recurring problem-solving patterns. A Relation Analyzer and a Relation Builder identify reusable relations and construct this hierarchy through \emph{Downward Atomic Extraction}, which captures minimal reusable capabilities, and \emph{Upward Abstract Induction}, which summarizes common problem-solving patterns. A task-driven self-evolution mechanism then folds each validated new skill back into the hierarchy by linking it to related skills, yielding a dynamic capability system that expands over time, reduces redundancy, and accumulates reusable knowledge.

Our contributions are as follows:
\begin{itemize}
    \item We propose \textsc{SkillPyramid}, a framework that organizes agent skills into a growing hierarchy and improves transferability through explicit reuse and dependency modeling.
    \item We design a pyramid organization mechanism with a Relation Analyzer and a Relation Builder, constructing atomic skills via downward extraction and abstract skills via upward induction.
    \item We introduce a task-driven self-evolution mechanism that retrieves, composes, and generates skills while incrementally updating \textsc{SkillPyramid}. Experiments on ALFWorld, WebShop, and ScienceWorld show 38.0\% higher average reward and 27.7\% fewer interaction steps across multiple backbone LLMs.
\end{itemize}

\section{Method}
\begin{figure*}[htbp]
    \centering
    \includegraphics[width=\textwidth]{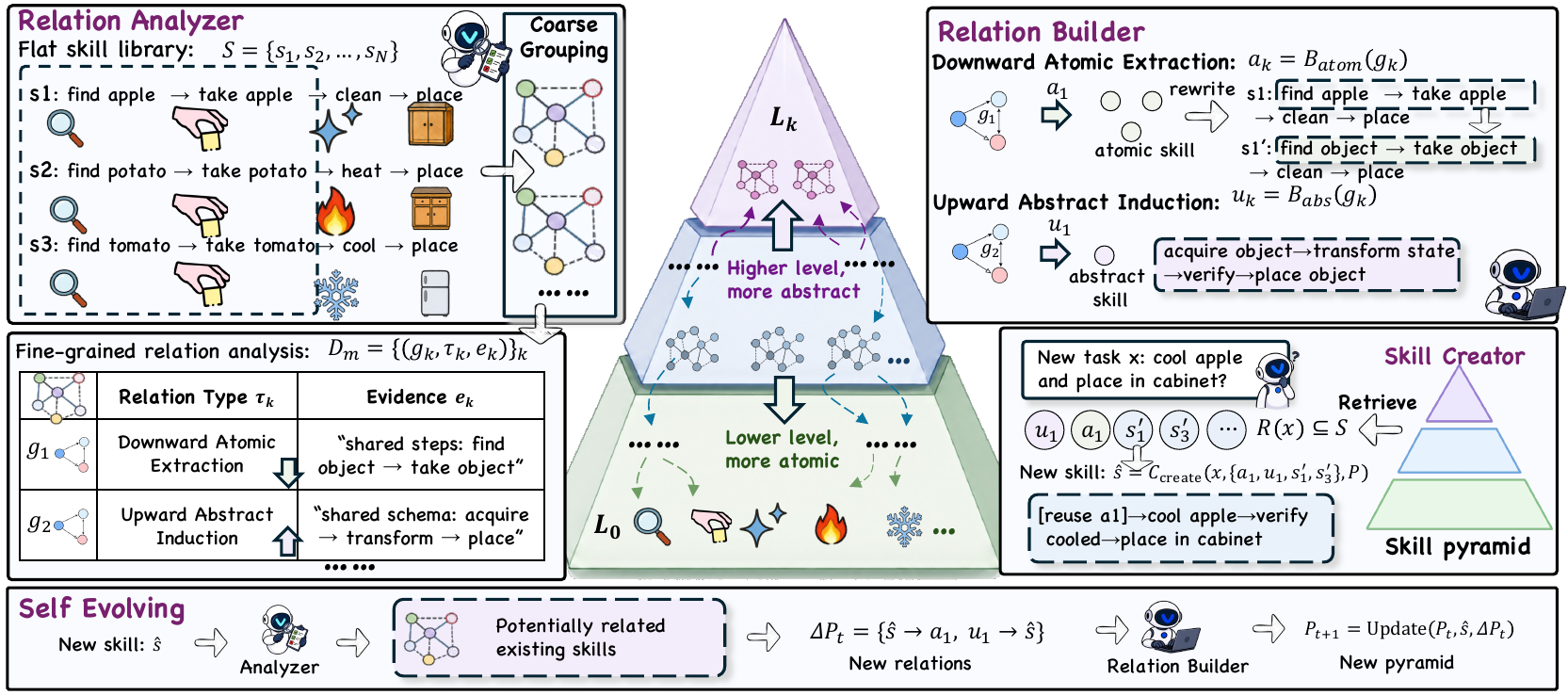}
    \caption{
    Overview of the proposed \textsc{SkillPyramid} framework. 
    The system analyzes an existing skill library, constructs downward atomic and upward abstract reuse relations, organizes skills into a hierarchical pyramid, creates skills from the pyramid, and incrementally evolves the pyramid as new skills are added.
    }
    \label{fig:pyramid}
\end{figure*}
Our method improves the generalization of a fixed skill collection by leveraging reusable patterns from existing high-quality skills. To this end, we construct a pyramidal, reusable, and continuously evolving skill system, where successful skills serve as reusable building blocks for related tasks rather than isolated modules. As illustrated in Figure~\ref{fig:pyramid}, The system is managed by a multi-agent framework that analyzes the existing skill library (\S\ref{sec:relation_analyzer}), constructs reuse relations among skills (\S\ref{sec:relation_builder}), synthesizes new skills grounded in the resulting hierarchy (\S\ref{sec:skill_creation}), and incrementally evolves as new skills are added (\S\ref{sec:self_evolution}).

\subsection{Problem Formalization}
\label{sec:formalization}
We consider an interactive agent setting where an agent receives a natural language task instruction $x$, interacts with an environment through actions, and receives task feedback after execution. At each step $t$, the agent observes the environment state $o_t$, takes an action $a_t$, and obtains a final reward $r \in [0,1]$ or a task completion signal after the episode ends. A complete task trajectory is denoted as $\tau=(x,o_0,a_0,o_1,\ldots,a_T,o_{T+1},r)$.

A skill $s$ is defined as a structured natural language program $s=(n,d,c)$, where $n$ is the skill name, $d$ is a short description, and $c$ is the skill content specifying its applicable conditions, execution procedure, and inputs/outputs.

Unlike methods that directly treat trajectories as training data for parameter updates, \textsc{SkillPyramid} converts task experience and existing skill descriptions into explicit, reusable skills.
Given a skill collection $\mathcal{S}=\{s_1,s_2,\ldots,s_N\}$, our goal is to construct and maintain a hierarchical \textsc{SkillPyramid} $\mathcal{P}=\{L_0,L_1,\ldots,L_K\}$ (illustrated in Figure~\ref{fig:pyramid}), where skills become increasingly abstract and compositional from the bottom layer $L_0$ to the top layer $L_K$. For a stream of tasks, \textsc{SkillPyramid} aims to improve task performance by reusing, composing, and evolving skills.

\subsection{\textsc{SkillPyramid} Construction}
\label{sec:construction}
To construct the pyramid, we employ two agents: a Relation Analyzer and a Relation Builder. The Analyzer identifies related skill groups, specifies the relation type for each, and dispatches the relation-construction tasks to the Builder. The Builder then extracts atomic skills, induces abstract skills, and rewrites the original skills. 
The Analyzer--Builder pipeline iterates until no new skill is produced or a maximum of $T$ rounds is reached, with each round's outputs feeding the next.

\subsubsection{Relation Analyzer}
\label{sec:relation_analyzer}
 
Given the current skill collection $\mathcal{S}=\{s_i\}_{i=1}^{N}$, directly analyzing all skills jointly may introduce substantial noise and lead to an excessive analysis workload. We therefore adopt a progressive analysis strategy that proceeds from coarse to fine granularity.
 
\paragraph{Coarse Grouping.}
Let $n_i$ and $d_i$ denote the name and description of skill $s_i$, respectively. The Relation Analyzer first performs a coarse grouping by inspecting only these two fields:
\begin{equation}
    \mathcal{G}^{0}
    =
    \mathcal{A}_{c}\!\left(\{(n_i,d_i)\}_{i=1}^{N}\right),
\end{equation}
where $\mathcal{A}_c$ denotes the \emph{coarse-grained analyzer} and $\mathcal{G}^{0}=\{G_1^{0},\ldots,G_M^{0}\}$ is the resulting set of coarse skill groups.
Each group $G_m^{0}\subseteq\mathcal{S}$ collects skills whose names and descriptions suggest potential relatedness in terms of functionality, task scope, or reuse potential. We use $G_m^{0}$ to refer interchangeably to this subset of skills and to its index set.

\paragraph{Fine-Grained Group Analysis.}
The Relation Analyzer then analyzes each coarse group $G_m^{0}$ in parallel, with tools to inspect the full structured content of every skill in the group. For each $G_m^{0}$, a \emph{fine-grained analyzer} $\mathcal{A}_f$ runs for at most $t$ rounds and produces a set of relation-construction assignments:
\begin{equation}
    \mathcal{D}_m = \mathcal{A}_f(G_m^{0}) = \{(g_k,\tau_k,e_k)\}_{k},
\end{equation}
where each $g_k \subseteq G_m^{0}$ is a fine-grained skill subset, $\tau_k$ specifies its relation type 
, and $e_k$ provides natural-language rationale and construction guidance. A single coarse group may yield multiple such subsets. The assignments from all coarse groups, $\mathcal{D}=\bigcup_m \mathcal{D}_m$, are then dispatched to Relation Builder for actual reuse-relation construction.
 
\subsubsection{Relation Builder}
\label{sec:relation_builder}
To capture reuse at both the execution level and the task-structure level, the Relation Builder operates in two complementary modes depending on $\tau_k$: \textit{Downward Atomic Extraction}, which surfaces shared low-level capabilities, and \textit{Upward Abstract Induction}, which summarizes shared high-level patterns. It takes the assignments produced by the Relation Analyzer as input and constructs the corresponding reuse relations among existing skills.
 
\paragraph{Downward Atomic Extraction.}
An atomic skill captures a minimal reusable capability that appears across multiple skills. Compared with concrete task skills, atomic skills are less tied to a specific task instance and can be reused across different environments or task categories.

For each assignment $(g_k,\tau_k,e_k)\in\mathcal{D}$ with $\tau_k=\textit{Downward Atomic Extraction}$, let $g_k=\{s_1,\ldots,s_m\}$. The Relation Builder extracts the common functional component shared by these skills as an atomic skill:
\begin{equation}
    a_k = B_{\mathrm{atom}}(g_k).
\end{equation}
Following the unified skill representation in \S\ref{sec:formalization}, $a_k$ takes the same form $(n_k, d_k, c_k)$ as any other skill, where $d_k$ summarizes the common goal of the shared capability.
To record explicit reuse relations introduced by the pyramid, we augment each skill with a \emph{dependency field} $\delta$ that stores references to other skills it reuses. After $a_k$ is extracted, each original skill $s_i\in g_k$ is rewritten to invoke $a_k$ through an explicit reuse reference:
\begin{equation}
    s_i' = \mathrm{Rewrite}\!\left(s_i,\rho(a_k)\right),\quad s_i\in g_k,
\end{equation}
where the reuse reference $\rho(a_k)=(n_k, q_k, w_k, o_k)$ records the reused skill's name $n_k$, a unique identifier $q_k$, the condition $w_k$ under which it should be reused, and the capability $o_k$ it provides. This reference is appended to the dependency field $\delta_i$ of $s_i'$ as a structured reuse statement.

If an existing skill is already fully covered by other skills, the Relation Builder directly treats this covered skill as the reusable module and updates the covering skills to reference it. To further reduce redundancy, highly similar extracted atomic skills are merged into a single canonical skill.

The extracted atomic skill $a_k$ is placed one layer below the source skills in the pyramid:
\begin{equation}
    \ell(a_k) = \min_{s\in g_k}\ell(s) - 1,
\end{equation}
where $\ell(\cdot)$ denotes the layer index of a skill.

\paragraph{Upward Abstract Induction.}
 
While atomic skills capture reusable low-level capabilities, a group of skills may also share similar high-level functional structures. For each assignment $(g_k,\tau_k,e_k)\in\mathcal{D}$ with $\tau_k=\textit{Upward Abstract Induction}$, the Relation Builder induces an abstract skill $u_k$ that summarizes the common solving pattern of $g_k$:
\begin{equation}
    u_k = B_{\mathrm{abs}}(g_k).
\end{equation}
The induced $u_k$ takes the same form $(n_k, d_k, c_k)$ as in \S\ref{sec:formalization}, with its dependency field $\delta_k$ recording references to the lower-level skills in $g_k$ that it composes. Unlike atomic skills, which focus on reusable operations, abstract skills focus on generalizable task structures: they describe how lower-level skills should be organized to solve a class of related tasks, rather than specifying only concrete execution steps.

The induced abstract skill $u_k$ is placed one layer above the source skills in the pyramid:
\begin{equation}
    \ell(u_k) = \max_{s\in g_k}\ell(s) + 1.
\end{equation}
In this way, $u_k$ serves as higher-level guidance, while the referenced concrete or atomic skills provide executable support for its realization.
 
\subsection{Skill Creation}
\label{sec:skill_creation}
While the pyramid built so far organizes existing skills into reusable structures, a new task often does not match any single skill exactly and instead requires composing several. To turn the pyramid from a passive repository into an active problem solver, we introduce a task-driven Skill Creator agent that generates new skills by composing relevant components from the existing hierarchy rather than synthesizing them from scratch.
Given a task instruction $x$, the Skill Creator retrieves relevant skills from the current \textsc{SkillPyramid}, analyzes how they can support the task, and synthesizes a new skill grounded in the existing hierarchy. Specifically, the Skill Creator retrieves a candidate set $\mathcal{R}(x)\subseteq\mathcal{S}$ by jointly considering task semantics, applicable conditions, and the constraints encoded in the pyramid.
 
Based on $\mathcal{R}(x)$, the Skill Creator generates a new skill in two stages:
(i) \textit{framework construction}, where higher-level (abstract) skills are prioritized to outline the task structure and high-level solving logic;
(ii) \textit{detail instantiation}, where lower-level (atomic) skills are prioritized to fill in executable operations, inputs, checks, and outputs.
If an existing skill fully covers the task, the Skill Creator directly reuses it. Otherwise, it composes a new task-specific skill:
\begin{equation}
    \hat{s} = C_{\mathrm{create}}\!\left(x,\mathcal{R}(x),\mathcal{P}\right),
\end{equation}
where $\mathcal{P}$ denotes the current pyramid, which encodes both the layer structure and the reuse relations among skills. The generated skill follows the same structured format as existing skills.
 
\subsection{Self-Evolution}
\label{sec:self_evolution}
A static skill library cannot keep pace with the open-ended stream of tasks that an agent encounters during deployment. To make \textsc{SkillPyramid} continually adaptive, we introduce an incremental self-evolution mechanism that absorbs each newly created skill into the existing hierarchy without rebuilding it from scratch.

At iteration $t$, the system maintains a skill collection $\mathcal{S}_t$ and a pyramid structure $\mathcal{P}_t$, which encodes both the layer assignment and the reuse relations among skills in $\mathcal{S}_t$. Given a newly created skill $\hat{s}$, the goal of self-evolution is to update $\mathcal{P}_t$ by connecting $\hat{s}$ with relevant prior skills. Before insertion, $\hat{s}$ is validated by the current task: successful candidates are refined from the execution trajectory and retained, while failed candidates are discarded.

Unlike the initial construction stage, the Relation Analyzer does not re-analyze the whole skill library. Instead, it treats $\hat{s}$ as a query skill, retrieves only potentially related existing skills, and dispatches the corresponding relation-construction tasks to the Relation Builder.

The Relation Builder then follows the same procedure as in \S\ref{sec:relation_builder} to construct new reuse relations involving $\hat{s}$. The resulting updates, denoted $\Delta\mathcal{P}_t$, are incrementally inserted into the pyramid:
\begin{equation}
    \mathcal{P}_{t+1} = \mathrm{Update}\!\left(\mathcal{P}_t,\,\hat{s},\,\Delta\mathcal{P}_t\right).
\end{equation}
This procedure allows \textsc{SkillPyramid} to absorb validated new skills incrementally while preserving its existing hierarchical organization.

\section{Experiment}
\label{sec:exp}

\subsection{Setup}
\paragraph{Benchmarks.}
We evaluate our method on three challenging benchmarks: ALFWorld~\citep{shridhar2021alfworld}, ScienceWorld~\citep{wang2022scienceworld}, and WebShop~\citep{yao2022webshop}. ALFWorld is a text-based environment that aligns embodied household tasks with their abstract textual counterparts. ScienceWorld is an interactive text environment in which agents must complete elementary-school-level science experiments. WebShop simulates a real-world e-commerce website in which agents are required to navigate, search, and purchase products. Further details on each benchmark are provided in Appendix~\ref{app:benchmarks}.

\paragraph{Baselines.}
We compare our approach against four baselines: ReAct~\citep{yao2022react}, Reflexion~\citep{shinn2023reflexion}, ExpeL~\citep{zhao2023expel}, and ReAct+Skills. ReAct interleaves chain-of-thought reasoning with environment actions. Reflexion and ExpeL are experience-based methods that reflect on past trajectories
to guide future decisions. The ReAct+Skills baseline and our proposed \textsc{SkillPyramid} method have access to the same initial skill library, ensuring a fair comparison. Further details on each baseline are provided in Appendix~\ref{app:baselines}.

\paragraph{Models.}
We evaluate all methods on a set of representative large language models, including DeepSeek-V3.2~\citep{deepseekai2025deepseekv32}, GPT-4.1~\citep{openai2025gpt41}, Gemini~2.5~Pro~\citep{comanici2025gemini25}, and Qwen3-235B~\citep{yang2025qwen3}. To ensure deterministic and reproducible outputs, the sampling temperature is set to $0$ for all models.
\begin{table*}[!t]
\centering

\setlength{\tabcolsep}{4pt}
\resizebox{\textwidth}{!}{%
\begin{tabular}{llcccccccccc}
\toprule
\multirow{3}{*}{Model} & \multirow{3}{*}{Method}
 & \multicolumn{4}{c}{ALFWorld} & \multicolumn{2}{c}{WebShop}
 & \multicolumn{4}{c}{ScienceWorld} \\
\cmidrule(lr){3-6} \cmidrule(lr){7-8} \cmidrule(lr){9-12}
 & & \multicolumn{2}{c}{Seen} & \multicolumn{2}{c}{Unseen} & \multicolumn{2}{c}{Seen}
   & \multicolumn{2}{c}{Seen} & \multicolumn{2}{c}{Unseen} \\
\cmidrule(lr){3-4}\cmidrule(lr){5-6}\cmidrule(lr){7-8}\cmidrule(lr){9-10}\cmidrule(lr){11-12}
 & & R$\uparrow$ & S$\downarrow$ & R$\uparrow$ & S$\downarrow$ & R$\uparrow$ & S$\downarrow$
     & R$\uparrow$ & S$\downarrow$ & R$\uparrow$ & S$\downarrow$ \\
\midrule
\multirow{5}{*}{DeepSeek-V3.2}
 & ReAct          & 66.4 & 19.5 & 69.4 & 19.3 & 31.6 & 24.1 & 69.9 & 17.6 & 64.7 & 19.3 \\
 & Reflexion      & 70.1 & 18.2 & 73.6 & 18.0 & \textbf{35.3} & \textbf{22.4} & 72.4 & 16.3 & 67.8 & 17.9 \\
 & ExpeL          & 67.9 & 18.9 & 76.1 & 17.4 & 29.2 & 24.0 & \underline{74.9} & \underline{16.0} & \underline{74.1} & 17.5 \\
 & ReAct + Skills & \underline{75.7} & \underline{15.6} & \underline{80.6} & \underline{14.7} & 27.0 & 25.1 & 71.6 & \textbf{15.9} & 68.5 & \textbf{16.6} \\
 & \textsc{SkillPyramid} (Ours) & \textbf{81.4} & \textbf{10.7} & \textbf{84.8} & \textbf{10.6} & \underline{32.0} & \underline{23.9} & \textbf{76.3} & 16.7 & \textbf{79.1} & \underline{17.0} \\
\midrule
\multirow{5}{*}{GPT-4.1}
 & ReAct          & 58.6 & 20.1 & 56.1 & 20.4 & 30.4 & 21.8 & 57.4 & 19.1 & 52.3 & 20.2 \\
 & Reflexion      & 64.2 & 18.5 & 62.8 & 18.6 & 34.6 & 20.1 & 66.8 & 17.0 & 61.6 & 18.2 \\
 & ExpeL          & 65.7 & 17.6 & 64.8 & 17.4 & 32.6 & \underline{18.7} & 72.1 & \textbf{14.4} & 66.8 & \underline{15.6} \\
 & ReAct + Skills & \underline{70.7} & \underline{16.4} & \underline{75.8} & \underline{16.2} & \underline{38.5} & 22.1 & \underline{78.6} & 17.6 & \underline{71.8} & 18.6 \\
 & \textsc{SkillPyramid} (Ours) & \textbf{77.1} & \textbf{15.1} & \textbf{85.1} & \textbf{13.1} & \textbf{42.5} & \textbf{15.3} & \textbf{86.6} & \underline{15.2} & \textbf{88.6} & \textbf{14.2} \\
\midrule
\multirow{5}{*}{Gemini~2.5~Pro}
 & ReAct          & 60.0 & 18.7 & 61.9 & 19.2 & 31.7 & 22.1 & 58.2 & 18.4 & 56.1 & 19.1 \\
 & Reflexion      & 67.2 & 17.3 & 69.4 & 17.7 & \textbf{38.2} & 19.9 & 66.8 & 16.9 & 63.2 & 17.5 \\
 & ExpeL          & 68.6 & 17.9 & 70.2 & 17.0 & 33.1 & 19.3 & 72.8 & \underline{15.0} & 67.4 & \underline{14.9} \\
 & ReAct + Skills & \textbf{75.0} & \underline{17.0} & \underline{73.9} & \underline{16.7} & 33.0 & \underline{18.7} & \underline{77.3} & 16.7 & \underline{79.1} & 17.0 \\
 & \textsc{SkillPyramid} (Ours) & \underline{73.6} & \textbf{16.0} & \textbf{80.6} & \textbf{11.9} & \underline{35.0} & \textbf{18.2} & \textbf{89.7} & \textbf{9.0} & \textbf{82.9} & \textbf{14.3} \\
\midrule
\multirow{5}{*}{Qwen3-235B}
 & ReAct          & 55.4 & 20.5 & 53.8 & 21.0 & 28.6 & 22.4 & 54.8 & 19.6 & 50.6 & 20.8 \\
 & Reflexion      & 61.8 & 19.0 & 60.2 & 19.5 & \underline{32.4} & 20.6 & 63.6 & 17.7 & 59.4 & 18.9 \\
 & ExpeL          & 63.2 & 18.2 & 62.4 & \underline{17.8} & 30.8 & \underline{19.2} & 69.4 & \underline{15.3} & 63.8 & 16.2 \\
 & ReAct + Skills & \underline{71.4} & \underline{17.3} & \underline{68.6} & 18.3 & 31.0 & 22.2 & \underline{74.2} & 15.5 & \underline{73.0} & \underline{15.9} \\
 & \textsc{SkillPyramid} (Ours) & \textbf{77.1} & \textbf{13.7} & \textbf{91.0} & \textbf{11.0} & \textbf{38.0} & \textbf{18.4} & \textbf{85.6} & \textbf{14.0} & \textbf{86.3} & \textbf{14.3} \\
\bottomrule
\end{tabular}%
}
\caption{Main results across ALFWorld, WebShop, and ScienceWorld. R denotes reward / success rate ($\uparrow$ higher is better); S denotes the average number of interaction steps ($\downarrow$ lower is better). The best result for each metric within each model is shown in \textbf{bold}, and the second-best is \underline{underlined}.}
\label{tab:main_results}
\end{table*}

\subsection{Main Results}

We conduct comprehensive experiments on ALFWorld, WebShop, and ScienceWorld with four representative backbone models. The overall results are summarized in Table~\ref{tab:main_results}. We report both task reward / success rate (R) and the average number of interaction steps (S), where higher reward and fewer steps indicate better agent performance.

\paragraph{\textsc{SkillPyramid} improves performance and generalization.}
Across different environments and backbone models, \textsc{SkillPyramid} achieves the best average performance among all evaluated methods. Averaged over all reward metrics in Table~\ref{tab:main_results}, \textsc{SkillPyramid} obtains an average reward of $73.7$, outperforming ReAct ($53.4$), Reflexion ($59.6$), ExpeL ($61.3$), and ReAct+Skills ($65.8$). Since ReAct+Skills and \textsc{SkillPyramid} are provided with the same initial skill library, this improvement shows that the gain does not merely come from access to additional skills, but from organizing skills into a reusable pyramidal structure. This advantage is especially clear on unseen tasks, where \textsc{SkillPyramid} achieves an average unseen-task reward of $84.8$, substantially higher than ReAct+Skills ($73.9$), ExpeL ($68.2$), Reflexion ($64.8$), and ReAct ($58.1$). These results suggest that the atomic and abstract skills induced by \textsc{SkillPyramid} capture transferable task-solving knowledge and enable stronger generalization to novel task instances.

\paragraph{\textsc{SkillPyramid} improves interaction efficiency.}
In addition to improving task reward, \textsc{SkillPyramid} also reduces the average number of interaction steps. Across all step metrics, \textsc{SkillPyramid} uses only $14.6$ steps on average, compared with $20.2$ for ReAct, $18.5$ for Reflexion, $17.4$ for ExpeL, and $17.7$ for ReAct+Skills. This efficiency gain is particularly evident on ALFWorld: for example, on DeepSeek-V3.2, \textsc{SkillPyramid} reduces the number of steps from $15.6$ to $10.7$ on seen tasks and from $14.7$ to $10.6$ on unseen tasks compared with ReAct+Skills. On Qwen3-235B, it further reduces the steps from $17.3$ to $13.7$ on ALFWorld seen tasks and from $18.3$ to $11.0$ on unseen tasks. These results indicate that hierarchical skill reuse helps agents reach successful states through shorter and more targeted action sequences.

\section{Analysis}
\label{sec:analyses}

In this section, we further analyze the contribution of each component in \textsc{SkillPyramid}. 
We focus on two questions: whether downward atomic extraction, upward abstract induction, and task-driven self-evolution are necessary for the final performance gain, and how self-evolution affects agent performance as the skill system is gradually updated.

\paragraph{Component Contribution.}

To evaluate the contribution of each component in \textsc{SkillPyramid}, we conduct ablation experiments on ALFWorld and ScienceWorld with DeepSeek-V3.2 as the backbone (Table~\ref{tab:ablation}). We consider four ablated variants: (i) \textit{w/o atomic skills}, in which Downward Atomic Extraction is disabled so that no atomic skills are produced and source skills retain their original form; (ii) \textit{w/o abstract skills}, in which Upward Abstract Induction is disabled so that no higher-level guidance is induced; (iii) \textit{w/o self-evolution}, in which the pyramid is frozen after initial construction and newly created skills are not folded back; and (iv) \textit{w/ scratch-generated skills}, in which the Skill Creator generates new skills without retrieving or reusing skills from the pyramid. As reference baselines, we also report vanilla \textsc{ReAct} without external skills and \textsc{ReAct w/ flat skills}, which uses the same initial skill collection but without pyramidal organization.
Removing atomic skills drops the average reward from $81.4\!\to\!78.6$ on ALFWorld and $76.3\!\to\!75.3$ on ScienceWorld, while removing abstract skills yields $81.4\!\to\!80.0$ and $76.3\!\to\!71.1$, respectively. This asymmetry suggests that atomic skills mainly help on benchmarks rich in reusable low-level operations, whereas abstract skills are more critical for multi-step procedural composition—a point we revisit in \S\ref{sec:discussion}. Freezing the pyramid further reduces performance to $76.4/73.7$, confirming that incremental incorporation of new skills is needed for sustained adaptation. Most strikingly, scratch-generation collapses the reward to $64.3/70.1$, falling below the flat-skill baseline ($75.7$ on ALFWorld) and indicating that ungrounded synthesis injects more noise than benefit. Overall, every component contributes, and grounding new skills in the existing hierarchy is essential for reliable skill synthesis.

\begin{table}[t]
\centering
\small
\resizebox{\columnwidth}{!}{
\begin{tabular}{@{}lcc@{}}
\toprule
\textbf{Method} & \textbf{ALFWorld} & \textbf{ScienceWorld} \\
\midrule
\multicolumn{3}{@{}l}{\textit{Baselines}} \\
\quad \textsc{ReAct} 
    & 66.4 & 69.9 \\
\quad \textsc{ReAct} w/ flat skills 
    & 75.7 & 71.6 \\

\addlinespace[2pt]
\multicolumn{3}{@{}l}{\textit{Ablations}} \\
\quad w/o Atomic Skills 
    & 78.6 & 75.3 \\
\quad w/o Abstract Skills 
    & 80.0 & 71.1 \\
\quad w/o Self-evolution 
    & 76.4 & 73.7 \\
\quad w/ Scratch-generated Skills 
    & 64.3 & 70.1 \\

\addlinespace[2pt]
\multicolumn{3}{@{}l}{\textit{Full method}} \\
\quad \textsc{SkillPyramid} 
    & \textbf{81.4} & \textbf{76.3} \\
\bottomrule
\end{tabular}
}
\caption{
Ablation results on ALFWorld and ScienceWorld using DeepSeek-V3.2 as the backbone model.
}
\label{tab:ablation}
\end{table}

\begin{figure}[t]
    \centering
    \includegraphics[width=\linewidth]{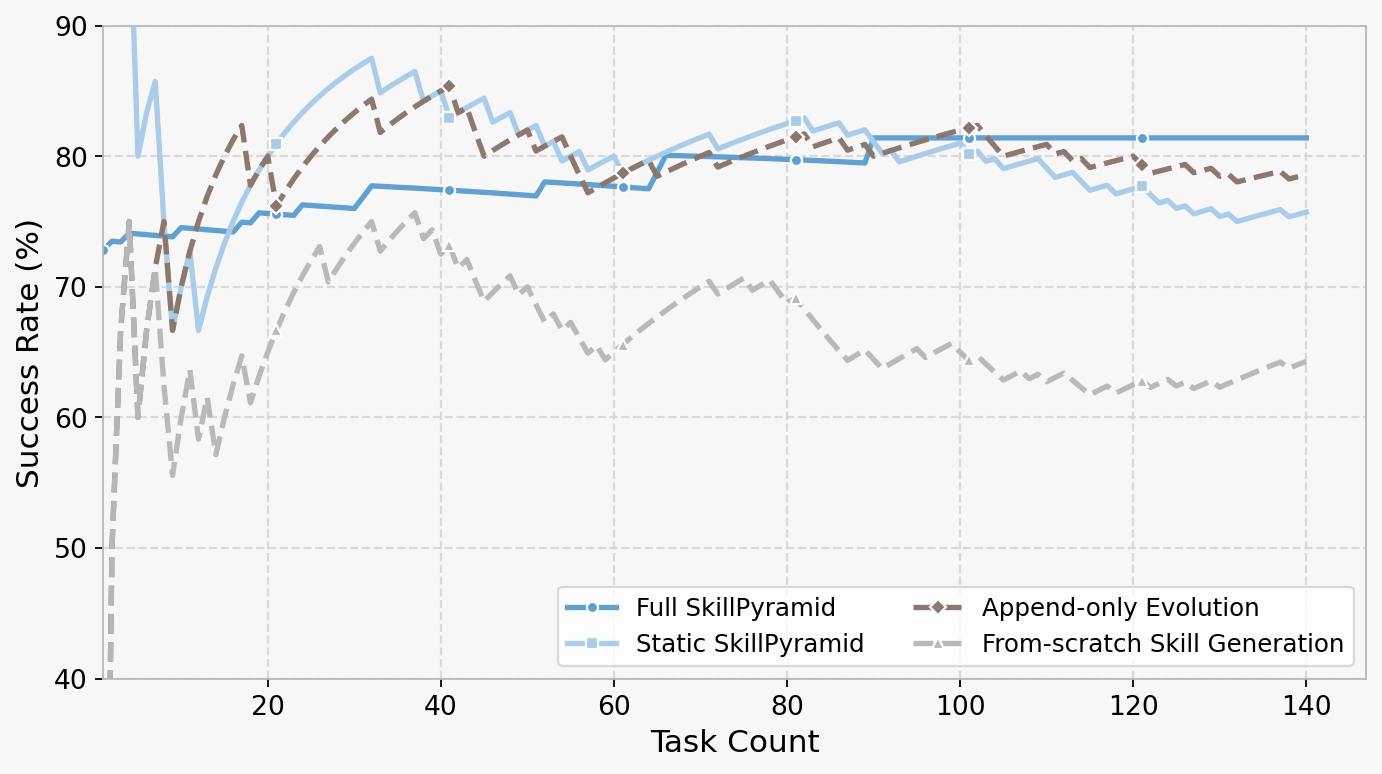}
    \caption{
    Self-evolution learning curves over incoming tasks on the ALFWorld seen split, obtained using the DeepSeek-V3.2.
    The full \textsc{SkillPyramid} maintains stronger and more stable late-stage performance than static, append-only, and from-scratch variants.
    }
    \label{fig:self_evolution_curve}
\end{figure}

\paragraph{Effect of Self-Evolution.}
We further analyze how self-evolution affects performance as the skill system is updated over incoming tasks.
Figure~\ref{fig:self_evolution_curve} compares the full \textsc{SkillPyramid} with three variants: a static version without skill updates, an append-only version without consolidation, and a from-scratch version that generates new skills without reusing the pyramid.
The full \textsc{SkillPyramid} maintains the most stable late-stage performance and gradually converges to a high success rate.
In contrast, the static variant performs well at early stages but degrades over time, suggesting that a fixed skill system cannot continuously adapt to new tasks.
Append-only evolution also achieves competitive performance in the middle stage, but its performance becomes less stable as redundant and weakly organized skills accumulate.
The from-scratch variant remains substantially worse, showing that grounding new skills in the existing pyramid is important for reliable skill generation.
Overall, these results indicate that self-evolution is most effective when newly generated skills are reused through the hierarchical pyramid rather than simply appended or generated independently.

\section{Discussion}
\label{sec:discussion}

The results suggest that the effectiveness of \textsc{SkillPyramid} comes not only from exposing agents to external skills, but from reorganizing these skills into reusable low-level components and high-level procedural guidance. 
We further discuss two questions: how atomic and abstract skills contribute differently, and whether hierarchical consolidation remains useful when scaling to noisy web-mined skills.

\paragraph{Why Do Atomic and Abstract Skills Help?}
To avoid post-hoc grouping, we partition tasks \emph{before} the ablations using action statistics from oracle trajectories. For each task $x$, define the low-level action mass
\[
m_{\mathrm{low}}(x)=\tfrac{1}{T_x}\sum_{t=1}^{T_x}\mathbb{I}[a_t\in\mathcal{A}_{\mathrm{low}}],
\]
where $\mathcal{A}_{\mathrm{low}}$ collects reusable primitives (\textit{navigation}, \textit{search}, \textit{pick}, \textit{put}, \textit{open}, \textit{use}, \textit{observe}, \textit{measure}), and let $\ell_{\mathrm{dep}}(x)$ be the longest ordered dependency chain over state-changing or diagnostic motifs (e.g., \textit{clean-then-place}, \textit{heat-then-place}, \textit{test-then-observe}, \textit{measure-then-compare}). Using the empirical quartiles $Q_1,Q_3$ of each statistic over the evaluation set, we label $x$ as \emph{atomic-heavy} if $m_{\mathrm{low}}(x)\!\ge\!Q_3(m_{\mathrm{low}})$ and $\ell_{\mathrm{dep}}(x)\!\le\!Q_1(\ell_{\mathrm{dep}})$, and as \emph{abstract-heavy} if $m_{\mathrm{low}}(x)\!\le\!Q_1(m_{\mathrm{low}})$ and $\ell_{\mathrm{dep}}(x)\!\ge\!Q_3(\ell_{\mathrm{dep}})$; tasks in neither extreme are discarded as ambiguous. Table~\ref{tab:atomic_abstract_discussion} shows an asymmetric pattern: removing atomic skills degrades atomic-heavy tasks more, whereas removing abstract skills disproportionately hurts abstract-heavy tasks. This supports our hypothesis that atomic skills primarily provide reusable execution, while abstract skills drive high-level decomposition and the composition of lower-level skills.

\begin{table}[t]
\centering
\small
\setlength{\tabcolsep}{3.5pt}
\resizebox{\columnwidth}{!}{
\begin{tabular}{lcccccc}
\toprule
\textbf{Task Group} 
& \textbf{\#Tasks} 
& $\boldsymbol{m_{\mathrm{low}}}$ 
& $\boldsymbol{\ell_{\mathrm{dep}}}$ 
& \textbf{Full} 
& \textbf{w/o Atomic} 
& \textbf{w/o Abstract} \\
\midrule
Atomic-heavy   
& 184 
& 0.93 
& 1.96 
& \textbf{82.6} 
& 76.1 {\scriptsize ($-$6.5)} 
& 80.2 {\scriptsize ($-$2.4)} \\
Abstract-heavy 
& 147 
& 0.43
& 5.14 
& \textbf{78.9} 
& 76.8 {\scriptsize ($-$2.1)} 
& 70.9 {\scriptsize ($-$8.0)} \\
\bottomrule
\end{tabular}
}
\caption{
Action-based task grouping and ablation results.
$m_{\mathrm{low}}$ and $\ell_{\mathrm{dep}}$ denote atomicity and dependency depth.
Parentheses indicate drops from full \textsc{SkillPyramid}.
}
\label{tab:atomic_abstract_discussion}
\end{table}

\paragraph{Pyramidal Consolidation Improves Web-Mined Skills.}
To examine whether \textsc{SkillPyramid} remains useful beyond benchmark-specific skill libraries, we evaluate web-mined skills on GAIA-Lite, a lightweight 75-task subset sampled from the public validation split of GAIA~\citep{mialon2023gaia}. Starting from 150K web-mined candidate skills, we retain 2,271 high-quality skills and compare three settings: no web skills, flat web skills, and the same web skills organized by \textsc{SkillPyramid}. As shown in Table~\ref{tab:web_skill_task_performance}, flat web skills already improve over the no-skill baseline, while \textsc{SkillPyramid} Web Skills further achieve higher accuracy and fewer execution steps on both Level 1 and Level 2 tasks. This indicates that the gain comes not only from acquiring more skills, but from consolidating noisy web-mined skills into a pyramidal structure with atomic operations, abstract task schemas, and explicit reuse relations.

\begin{table}[t]
\centering
\small
\setlength{\tabcolsep}{4pt}
\resizebox{\columnwidth}{!}{
\begin{tabular}{lccc}
\toprule
\textbf{Setting} &
\textbf{Level 1} &
\textbf{Level 2} &
\textbf{Overall} \\
&
\textbf{Acc.$\uparrow$ / Step$\downarrow$} &
\textbf{Acc.$\uparrow$ / Step$\downarrow$} &
\textbf{Acc.$\uparrow$ / Step$\downarrow$} \\
\midrule
No Web Skills
& 57.8 / 5.6
& 43.3 / \textbf{4.7}
& 52.0 / 5.2 \\
Flat Web Skills
& 62.2 / 5.3
& 46.7 / 6.8
& 56.0 / 5.9 \\
\textsc{SkillPyramid} Web Skills
& \textbf{68.9} / \textbf{4.5}
& \textbf{53.3} / 5.3
& \textbf{62.7} / \textbf{4.8} \\
\bottomrule
\end{tabular}
}
\caption{
Performance of web-mined skills on GAIA-Lite, a 75-task subset of GAIA covering Level 1 and Level 2 tasks.
Acc. denotes exact-match accuracy, and Step denotes the average number of steps.
}
\label{tab:web_skill_task_performance}
\end{table}

\section{Related Work}

\paragraph{LLM agents and experience reuse.}
LLM agents have made substantial progress on long-horizon interactive tasks by interleaving reasoning, action, and experience reuse~\citep{yao2022react,shinn2023reflexion,zhao2023expel,schick2023toolformer,qin2023toolllm}. Existing methods can be broadly grouped into reasoning--action coupling, experience-driven self-improvement, and tool-augmented orchestration. ReAct interleaves verbal reasoning with environment actions to support sequential decision making~\citep{yao2022react}. Building on this paradigm, Reflexion accumulates verbal self-reflection across episodes to refine future trials~\citep{shinn2023reflexion}, while ExpeL consolidates cross-task experiential insights to bootstrap unseen tasks~\citep{zhao2023expel}. Beyond pure reasoning, another line augments agents with external tools and expert modules: MRKL connects language models with symbolic reasoners through modular routing~\citep{karpas2022mrkl}, and Toolformer learns API invocation from self-supervised demonstrations~\citep{schick2023toolformer}. 
Subsequent work scales this paradigm through expert-model orchestration~\citep{shen2023hugginggpt}, plug-and-play tool composition~\citep{lu2023chameleon}, and large-scale API use~\citep{qin2023toolllm}.

\paragraph{Skill organization and evolution.}
Reusable skills have emerged as an essential substrate for long-horizon agent learning, allowing agents to consolidate interaction experience into transferable capabilities~\citep{liang2022codepolicies,wang2023voyager,wang2026skillx,mi2026skillpro,fang2025memp}. Existing work can be broadly categorized into executable skill libraries, procedural memory accumulation, and structured skill knowledge bases. Code as Policies synthesizes executable policy code by composing low-level control primitives for embodied control~\citep{liang2022codepolicies}. DEPS plans in open worlds through description, explanation, planning, and selection~\citep{wang2023deps}, and Voyager builds an ever-growing executable skill library for lifelong Minecraft exploration~\citep{wang2023voyager}. From a procedural memory perspective, Memp explores reusable procedural traces accumulated from agent interaction~\citep{fang2025memp}, while Skill-Pro learns reusable skills via non-parametric PPO~\citep{mi2026skillpro}. More recently, structured skill knowledge bases have emerged: SkillNet creates, evaluates, and connects skills under a unified interface~\citep{liang2026skillnet}, and SkillX automatically constructs multi-level skill knowledge bases from agent trajectories~\citep{wang2026skillx}.

\paragraph{Feedback, evidence, and multimodal reasoning.}
Complementary work studies how supervision and context affect model capabilities. OPRD distills reminder-conditioned guidance into long-horizon coding agents~\citep{luo2026oprdonpolicyreminderdistillation}, while SWE-Nexus constructs and composes software engineering tasks from test execution~\citep{yu2026swenexusautomatedcompositionsoftware}. Beyond coding, TabSieve explicitly selects in-table evidence for prediction~\citep{wang2026tabsieveexplicitintableevidence}, and stable adaptive thinking uses advantage shaping and length-aware gradient regulation to balance reasoning accuracy and efficiency~\citep{xu2026stableadaptivethinkingadvantage}. In multimodal settings, Seeing with You coordinates perception and reasoning through cooperative Observer and Solver roles~\citep{miao2026seeingyouperceptionreasoningcoevolution}, and Re$^3$Cap uses retrieval-guided refinement and reinforcement learning for image captioning~\citep{jia2026re3capretrievalguidedrefinementimage}. In autonomous driving, AdaThinkDrive uses reinforcement learning to adaptively select between fast answering and slow chain-of-thought reasoning~\citep{luo2025adathinkdrive}, while ELF-VLA uses structured failure diagnostics to generate refined samples for reinforcement learning~\citep{luo2026elfvla}. Context can also introduce vulnerabilities: Visual Contextual Attack studies image-driven context injection for jailbreaking multimodal models~\citep{miao2025visualcontextualattackjailbreaking}. These studies examine supervision, policy optimization, evidence use, and contextual vulnerabilities rather than the organization of reusable skills. \textsc{SkillPyramid} instead consolidates skills into an evolving hierarchy without updating model parameters; extending this hierarchy to multimodal settings remains future work.

\section{Conclusion}

We presented \textsc{SkillPyramid}, a hierarchical skill consolidation framework for improving the reuse, transfer, and evolution of agent skills. By analyzing an existing skill collection, \textsc{SkillPyramid} extracts atomic reusable skills, induces abstract guidance skills, and maintains explicit reuse and dependency relations among them. Its task-driven self-evolution mechanism further enables agents to compose, validate, and incorporate new skills during execution. Experiments on ALFWorld, WebShop, and ScienceWorld show that \textsc{SkillPyramid} consistently improves task reward while reducing interaction steps on average across multiple LLMs, suggesting that structured skill consolidation is a promising direction for building agents that accumulate reusable capabilities over time.

\section*{Limitations}
This paper has several limitations. First, the initial pyramid construction requires a one-time analysis over the full skill collection, and scaling to substantially larger repositories may require more efficient construction strategies. Second, \textsc{SkillPyramid} still depends on sufficiently capable construction models and informative skill descriptions; performance degrades with weaker backbones or highly ambiguous skills. Third, success-gated validation does not guarantee semantic correctness, and the current framework lacks a hard rollback mechanism for removing previously admitted but later found erroneous skills, which may propagate through subsequent reuse. Fourth, our evaluation is limited to text-based interactive benchmarks, leaving multimodal settings unexplored. Finally, the pyramid models only two reuse axes and does not capture richer temporal or causal relations.

\section*{Ethical Considerations}
Our experiments use synthetic interactive benchmarks (ALFWorld, ScienceWorld, WebShop, and GAIA-Lite) that contain no personally identifying or offensive content, and involve no human subjects. As a general framework for skill accumulation and self-evolution, \textsc{SkillPyramid} nevertheless inherits the dual-use concerns of capable LLM agents: a continually evolving skill library could, in principle, accumulate skills aligned with harmful objectives if deployed without task-level safeguards. We recommend that downstream users couple \textsc{SkillPyramid} with task-level access control and skill-level auditing, and apply dedicated content review to web-mined skill libraries, before deployment in user-facing applications.

\paragraph{Use of AI Assistants.} AI assistants were used to aid coding and to polish the writing of this paper. All scientific contributions, including the framework design, experimental setup, analyses, and conclusions, are the authors' own.

\section*{Acknowledgments}
This work was supported by the Beijing Major Science and Technology Project (No. Z251100008125025), the National Natural Science Foundation of China (No.62376270) and the independent research project of the Key Laboratory of Cognition and Decision Intelligence for Complex Systems. This work  was supported  by Beijing Academy of Artificial Intelligence (BAAI).

\bibliography{custom}

\appendix

\section{Experimental Setup and Implementation Details}
\label{sec:appendix}

\subsection{Benchmarks Implementation}
\label{app:benchmarks}
\paragraph{ALFWorld.}
ALFWorld~\cite{shridhar2021alfworld} is an interactive household benchmark that aligns abstract TextWorld games with embodied ALFRED tasks, requiring agents to complete long-horizon household goals by navigating rooms, observing state changes, and manipulating objects through textual actions. Following prior language-agent evaluations~\cite{yao2022react,shinn2023reflexion} and the SkillNet protocol~\cite{liang2026skillnet}, we use \texttt{eval\_in\_distribution} as the seen split and \texttt{eval\_out\_of\_distribution} as the unseen split, which contain 140 and 134 tasks, respectively.

\paragraph{ScienceWorld.}
ScienceWorld~\cite{wang2022scienceworld} is an interactive text-based benchmark for evaluating grounded scientific reasoning, where agents must perform experiments, operate instruments, and reason about changing world states in environments inspired by elementary science curricula. Following SkillNet~\cite{liang2026skillnet}, we use the dev split defined by \texttt{src/scienceworld/data/valid\_indices.json} as the seen split and \texttt{src/scienceworld/data/test\_indices.json} as the unseen split; these splits contain 194 and 211 task variations, respectively, covering 24 ScienceWorld task types.

\paragraph{WebShop.}
WebShop~\cite{yao2022webshop} is a simulated e-commerce web environment with real-world products and crowd-sourced shopping instructions, where agents search, compare, customize, and purchase products according to natural-language user goals. Following SkillNet~\cite{liang2026skillnet}, we evaluate on a fixed set of 200 randomly sampled WebShop tasks specified by \texttt{src/webshop/data/test\_indices.json}.

\subsection{Baselines Implementation}
\label{app:baselines}

\paragraph{ReAct.}
ReAct \citep{yao2022react} interleaves language reasoning with executable actions. Following the configuration of \citet{liang2026skillnet}, we implement a ReAct-style loop in which the agent observes the current state, produces a \texttt{Thought} and a valid \texttt{Action}, and updates its context with environment feedback. ReAct uses no external skills, reflections, or experience memory, and shares the same prompts as our evaluation setting.

\paragraph{Reflexion.}
Reflexion \citep{shinn2023reflexion} extends ReAct with verbal self-reflection. After each failed or low-reward trial, the model summarizes the error and stores a concise reflection for future attempts. These reflections are used only as textual guidance and are not transformed into structured skills or reuse relations.

\paragraph{ExpeL.}
ExpeL \citep{zhao2023expel} distills past trajectories into reusable natural-language lessons. We extract general strategies and failure patterns from previous tasks and provide the retrieved lessons as additional context for a ReAct-style agent. Unlike SkillPyramid, ExpeL keeps a flat experience memory without atomic extraction, abstract induction, or hierarchical updating.

\paragraph{ReAct+Skills.}
ReAct+Skills equips ReAct with the same initial skill library used by SkillPyramid. For each task, relevant skills are retrieved and inserted into the system prompt. This baseline isolates the effect of skill access, as it uses a flat skill library without hierarchical consolidation, relation construction, or self-evolution.

\subsection{Additional Baseline Comparisons}
\label{app:additional_baselines}

\paragraph{Comparison with SkillNet and SkillX.}
We further compare \textsc{SkillPyramid} with SkillNet~\cite{liang2026skillnet} and SkillX~\cite{wang2026skillx} on the ALFWorld unseen split using DeepSeek-V3.2. SkillNet constructs its skill library from the same training split as \textsc{SkillPyramid}, while SkillX derives its skills from trajectories collected on the same training split. All methods are evaluated on the identical unseen split.

\begin{table}[t]
\centering
\begin{tabular}{lcc}
\toprule
Method & R $\uparrow$ & S $\downarrow$ \\
\midrule
ReAct & 69.4 & 19.3 \\
ReAct + Skills & 80.6 & 14.7 \\
ReAct + SkillNet & 83.6 & 14.8 \\
ReAct + SkillX & 79.9 & 16.6 \\
ReAct + \textsc{SkillPyramid} & \textbf{84.8} & \textbf{10.6} \\
\bottomrule
\end{tabular}
\caption{Comparison with SkillNet and SkillX on the ALFWorld unseen split using DeepSeek-V3.2.}
\label{tab:skillnet_skillx}
\end{table}

\textsc{SkillPyramid} achieves 1.2 points higher reward than SkillNet while using 4.2 fewer interaction steps, and improves over SkillX by 4.9 reward points and 6.0 fewer steps.

\paragraph{Comparison with GPT-5.5 Codex.}
To examine whether \textsc{SkillPyramid} transfers beyond a ReAct-style agent, we conduct a controlled evaluation with GPT-5.5 Codex on the ALFWorld unseen split. Both settings use the same benchmark interface, evaluation split, and metrics.

\begin{table}[t]
\centering
\small
\begin{tabular}{lcc}
\toprule
Method & R $\uparrow$ & S $\downarrow$ \\
\midrule
GPT-5.5 Codex & 88.1 & 14.5 \\
GPT-5.5 Codex + \textsc{SkillPyramid} & \textbf{91.8} & \textbf{12.3} \\
\bottomrule
\end{tabular}
\caption{Controlled comparison with GPT-5.5 Codex on the ALFWorld unseen split.}
\label{tab:codex}
\end{table}
These results support the positioning of \textsc{SkillPyramid} as a skill-organization and reuse module rather than a mechanism restricted to the ReAct architecture.

\subsection{Transfer to an Embodied Multimodal Backbone}
\label{app:roboego}

We further evaluate \textsc{SkillPyramid} with the 7B RoboEgo model~\cite{yao2025roboegocardomnimodalmodel} on the ALFWorld unseen split, replacing only the backbone while keeping all other settings unchanged. As shown in Table~\ref{tab:roboego}, \textsc{SkillPyramid} improves reward from 27.6 to 29.9 (+2.3) and reduces interaction steps from 25.3 to 22.7 (-2.6), indicating consistent gains with an embodied multimodal backbone.

\begin{table}[t]
\centering
\begin{tabular}{lcc}
\toprule
Method & R $\uparrow$ & S $\downarrow$ \\
\midrule
ReAct + Skills & 27.6 & 25.3 \\
ReAct + \textsc{SkillPyramid} & \textbf{29.9} & \textbf{22.7} \\
\bottomrule
\end{tabular}
\caption{Transfer evaluation with RoboEgo on the ALFWorld unseen split.}
\label{tab:roboego}
\end{table}

\subsection{Additional Implementation Details}
During SkillPyramid construction, we use Qwen3.5-35B-A3B as the backbone model for all construction agents, and Qwen3-Embedding-4B to compute embedding similarities among skills for candidate retrieval, coarse grouping, and relation screening before full-content analysis. The initial skill library is constructed from the training splits of ALFWorld, ScienceWorld, and WebShop. The outer Analyzer--Builder pipeline is run for at most $3$ rounds, and the fine-grained analyzer within each Relation Analyzer call is run for at most $5$ rounds.
The initial skill libraries contain 65, 913, and 71 skills for ALFWorld, ScienceWorld, and WebShop, respectively. Each library is constructed exclusively from the training split of the corresponding benchmark, and the three benchmarks use separate skill libraries. During evaluation, each task retrieves skills from the complete pyramid associated with its benchmark, without manually pre-selecting skills for individual evaluation tasks. No evaluation task or trajectory is used during initial pyramid construction. Admitted skills are available only to subsequent tasks.

\subsection{End-to-End Construction Procedure}
\label{app:construction_procedure}

Algorithm~\ref{alg:skillpyramid} summarizes the end-to-end construction and self-evolution procedure of \textsc{SkillPyramid}.

\begin{algorithm}[t]
\caption{\textsc{SkillPyramid} Construction and Self-Evolution}
\label{alg:skillpyramid}
\begin{algorithmic}[1]
\Require Initial skill library $\mathcal{S}_0$, task stream $\mathcal{X}$
\Ensure Evolving skill pyramid $\mathcal{P}$

\State Initialize $\mathcal{P}$ with $\mathcal{S}_0$
\For{$r=1,\ldots,T$}
    \State Group potentially related skills by names, descriptions, and embedding similarity
    \State Relation Analyzer identifies fine-grained relation assignments
    \State Relation Builder performs atomic extraction or abstract induction
    \State Validate the constructed relations and update $\mathcal{P}$
    \If{no new skill or relation is produced}
        \State \textbf{break}
    \EndIf
\EndFor

\For{task $x \in \mathcal{X}$}
    \State Retrieve relevant skills $\mathcal{R}(x)$ from $\mathcal{P}$
    \State Generate a candidate skill $\hat{s}$ grounded in $\mathcal{R}(x)$
    \State Execute task $x$ using $\hat{s}$
    \If{task $x$ succeeds}
        \State Refine $\hat{s}$ from the successful trajectory
        \State Construct relations between $\hat{s}$ and related existing skills
        \State Validate and insert $\hat{s}$ into $\mathcal{P}$
    \Else
        \State Discard $\hat{s}$
    \EndIf
\EndFor

\State \Return $\mathcal{P}$
\end{algorithmic}
\end{algorithm}
\subsection{Construction Quality and Structural Validation}
\label{app:construction_quality}

\paragraph{Construction-quality audit.}
We manually evaluate sampled outputs from the ALFWorld skill library. A relation or generated skill is marked erroneous when it is unsupported by the source skills, overgeneralizes the shared capability, or is inserted at an incorrect hierarchy level.

\begin{table}[t]
\centering
\small
\begin{tabular}{lcc}
\toprule
Audit Item & \# Sampled & Error Rate $\downarrow$ \\
\midrule
Relation identification & 50 & 2.0\% \\
Atomic skill extraction & 20 & 10.0\% \\
Abstract skill induction & 20 & 20.0\% \\
New skill insertion & 30 & 13.3\% \\
\bottomrule
\end{tabular}
\caption{Manual construction-quality audit on sampled outputs from the ALFWorld skill library.}
\label{tab:construction_audit}
\end{table}

Relation identification reaches 98.0\% accuracy in this sample. Abstract skill induction has the highest error rate, reflecting the greater subjectivity of identifying shared high-level solving patterns. Given the limited sample size, we treat these results as a sampled diagnostic rather than a precise estimate of the overall construction error rate. Incorrect atomic or abstract skills may also propagate through subsequent reuse, making construction quality important for reliable long-term evolution.

\paragraph{Structural validation.}
After each skill insertion, \textsc{SkillPyramid} checks for incorrect edges, hierarchy inconsistencies, invalid cross-layer relations, and self-loops. Reuse edges are constrained to follow the hierarchy from higher-level skills to lower-level dependencies, which also prevents multi-node cycles. MERGE is applied only when skills are near-equivalent in their triggers, inputs, workflows, and outputs; skills with conflicting applicability conditions or workflows remain separate.
\subsection{Run-to-Run Stability}
\label{app:stability}

We conduct three independent runs using DeepSeek-V3.2 with temperature $0$. The runs use different task orders, independent API executions, and retrieval randomness. Table~\ref{tab:stability} reports the mean and standard deviation of reward $R$ and average interaction steps $S$.

\begin{table}[t]
\centering
\small
\resizebox{\columnwidth}{!}{
\begin{tabular}{llccc}
\toprule
Benchmark & Split & Method & R $\uparrow$ & S $\downarrow$ \\
\midrule
ALFWorld & Seen
& ReAct+Skills
& $75.4 \pm 1.2$
& $15.8 \pm 0.5$ \\

ALFWorld & Seen
& \textsc{SkillPyramid}
& $\mathbf{81.1 \pm 1.0}$
& $\mathbf{10.9 \pm 0.4}$ \\

ALFWorld & Unseen
& ReAct+Skills
& $80.1 \pm 1.5$
& $14.9 \pm 0.6$ \\

ALFWorld & Unseen
& \textsc{SkillPyramid}
& $\mathbf{84.5 \pm 1.3}$
& $\mathbf{10.8 \pm 0.5}$ \\

WebShop & Seen
& ReAct+Skills
& $27.4 \pm 1.6$
& $25.0 \pm 0.8$ \\

WebShop & Seen
& \textsc{SkillPyramid}
& $\mathbf{32.3 \pm 1.9}$
& $\mathbf{23.7 \pm 0.7}$ \\

ScienceWorld & Seen
& ReAct+Skills
& $71.9 \pm 1.4$
& $\mathbf{16.0 \pm 0.5}$ \\

ScienceWorld & Seen
& \textsc{SkillPyramid}
& $\mathbf{76.5 \pm 1.1}$
& $16.5 \pm 0.6$ \\

ScienceWorld & Unseen
& ReAct+Skills
& $68.2 \pm 1.8$
& $\mathbf{16.7 \pm 0.7}$ \\

ScienceWorld & Unseen
& \textsc{SkillPyramid}
& $\mathbf{78.8 \pm 1.5}$
& $17.1 \pm 0.6$ \\
\bottomrule
\end{tabular}
}
\caption{Run-to-run stability over three independent runs using DeepSeek-V3.2.}
\label{tab:stability}
\end{table}

\textsc{SkillPyramid} maintains consistent reward improvements across all five settings. The WebShop improvement is smaller and more variable, while ALFWorld and ScienceWorld show clearer reward gains. On ScienceWorld, the higher reward is accompanied by slightly more interaction steps, so the method does not universally reduce interaction length in every setting.

\subsection{Robustness Analysis}
\label{app:robustness}

We further examine the robustness of \textsc{SkillPyramid} to prompt perturbations, weaker construction backbones, and ambiguous input skill descriptions. All downstream evaluations use DeepSeek-V3.2 with temperature $0$.

\paragraph{Prompt perturbations.}
We compare the original Relation Analyzer and Relation Builder prompts with three meaning-preserving variants. \textit{Paraphrased} changes the wording, \textit{Order-perturbed} changes the order of rules and output instructions, and \textit{Instruction-perturbed} modifies explanatory language while preserving the core objective and output schema.

\begin{table}[t]
\centering
\small
\resizebox{\columnwidth}{!}{
\begin{tabular}{lccc}
\toprule
Prompt Variant
& ALFWorld Unseen
& ScienceWorld Unseen
& WebShop \\
\midrule
Original & 84.8 & 79.1 & 32.0 \\
Paraphrased & 80.1 & 78.4 & 31.0 \\
Order-perturbed & 82.4 & 77.8 & 31.5 \\
Instruction-perturbed & 78.7 & 80.2 & 30.0 \\
\bottomrule
\end{tabular}
}
\caption{Reward under prompt perturbations.}
\label{tab:prompt_robustness}
\end{table}

ScienceWorld and WebShop remain relatively stable under these perturbations, whereas ALFWorld is more sensitive to prompt wording, with a maximum reward drop of 6.1 points. We therefore do not claim prompt invariance.

\paragraph{Weaker construction backbones.}
We keep DeepSeek-V3.2 fixed for downstream evaluation and vary only the backbone used for pyramid construction.

\begin{table}[t]
\centering
\resizebox{\columnwidth}{!}{
\begin{tabular}{lccc}
\toprule
Construction Backbone
& ALFWorld Unseen
& ScienceWorld Unseen
& WebShop \\
\midrule
ReAct+Skills, no pyramid & 80.6 & 68.5 & 27.0 \\
Qwen3.5-35B-A3B & 84.8 & 79.1 & 32.0 \\
Qwen3-14B & 82.1 & 77.4 & 31.0 \\
Qwen2.5-7B-Instruct & 82.8 & 74.2 & 29.5 \\
Qwen2.5-3B-Instruct & 76.7 & 70.8 & 27.0 \\
\bottomrule
\end{tabular}
}
\caption{Reward with different pyramid-construction backbones.}
\label{tab:backbone_robustness}
\end{table}

Qwen3-14B and Qwen2.5-7B-Instruct preserve most of the gains, whereas Qwen2.5-3B-Instruct loses the ALFWorld improvement and only matches the flat baseline on WebShop. These results indicate that \textsc{SkillPyramid} does not require the strongest construction model, but does require sufficient relation-analysis and abstract-induction capability.

\paragraph{Ambiguous skill descriptions.}
We progressively remove information from the initial skills. \textit{Object-masked} replaces concrete object names with generic placeholders; \textit{Description-ambiguous} generalizes the skill name and description while retaining the main content; and \textit{Highly ambiguous} additionally removes part of the skill content, leaving only the high-level goal and coarse steps.

\begin{table}[t]
\centering
\resizebox{\columnwidth}{!}{
\begin{tabular}{lccc}
\toprule
Input Skill Description
& ALFWorld Unseen
& ScienceWorld Unseen
& WebShop \\
\midrule
ReAct+Skills, no pyramid & 80.6 & 68.5 & 27.0 \\
Original & 84.8 & 79.1 & 32.0 \\
Object-masked & 82.1 & 77.8 & 31.0 \\
Description-ambiguous & 82.8 & 75.3 & 30.5 \\
Highly ambiguous & 78.4 & 70.6 & 27.5 \\
\bottomrule
\end{tabular}
}
\caption{Reward under increasingly ambiguous input skill descriptions.}
\label{tab:ambiguity_robustness}
\end{table}

Partial information loss causes moderate degradation while retaining most of the benefit. Under highly ambiguous descriptions, performance drops substantially, and ALFWorld falls below the flat-skill baseline. This shows that \textsc{SkillPyramid} depends on sufficiently informative input skills and cannot reliably recover reusable relations from severely underspecified descriptions.

\subsection{WebShop Failure Analysis}
\label{app:webshop_failure}

Our inspection of failed WebShop trajectories reveals two recurring failure modes.

\paragraph{Fine-grained option errors.}
The agent may follow the correct search, browse, and purchase procedure but still select the wrong size, color, quantity, price range, or product variant. These local constraints often determine task success but are not fully captured by a high-level task structure.

\paragraph{Overgeneralized search guidance.}
Upward Abstract Induction may remove rare but essential lexical attributes from the original instruction. The resulting strategy can remain reasonable at a high level while retrieving a similar product that does not satisfy all constraints.

Unlike ALFWorld and ScienceWorld, where tasks often contain reusable procedural structures, WebShop has relatively shallow high-level interaction procedures and relies heavily on instance-specific attribute matching. Abstraction therefore provides less guidance in this setting and can occasionally overgeneralize.

\subsection{Compute and Model Budget}
All experiments use API calls only and no model parameters were fine-tuned, updated, or trained. The main evaluation spans 17,580 episodes (879 instances × 5 methods × 4 backbones). Total API usage: ~8.4K calls (21.6M input / 3.1M output tokens) for the one-time SkillPyramid construction, ~315K calls (0.92B / 46.8M tokens) for the main evaluation, and ~24K calls (71.5M / 5.4M tokens) for ablations and analysis. Local computation (environment simulation, retrieval, indexing, logging) ran on CPU-only nodes. 

\subsection{Prompt Templates}
\label{app:temp}

\paragraph{Prompt Template for Relation Analyzer}
The Relation Analyzer identifies reusable relations in the skill library. Its prompt follows a two-stage process: \textsc{Screen} filters candidate groups using names, descriptions, and similarity signals, while \textsc{Decide} inspects full \texttt{SKILL.md} files when needed and outputs relation-construction assignments for the Relation Builder.
\begin{verifierbox}{Relation Analyzer Prompt}
\begin{lstlisting}[basicstyle=\ttfamily\small, breaklines=true]
You are the RELATION ANALYZER in SKILLPYRAMID. You may be called multiple times.
Each time you are called, you will be clearly informed that you are currently in one of the following two stages.

STAGES
- SCREEN: Inspect skill names, descriptions and similarity signals to decide which candidate pairs or groups deserve DECIDE-stage analysis.
- DECIDE: Produce the final relation assignment for the RELATION BUILDER. Use the available tools to inspect full `SKILL.md` contents when necessary.

Your job is to identify only relations that are useful for building a hierarchical reusable skill pyramid. Be conservative and evidence-grounded.

INPUT FOR SCREEN
{
  "stage": "SCREEN",
  "candidates": [
    {
      "skills": [
        {"name": "<skill name>", "description": "<skill description>"}
      ],
      "similarity": "<optional score or signal>"
    }
  ]
}

INPUT FOR DECIDE
{
  "stage": "DECIDE",
  "candidate": {
    "skills": [...],
    "screen_reason": "<optional reason>",
    "similarity": "<optional score or signal>"
  }
}

TOP-LEVEL ASSIGNMENTS

DOWNWARD_ATOMIC_EXTRACTION: Resolve the relation by creating, reusing, or canonicalizing a lower-level atomic capability. It includes three relation types:
- SHARED_PART: Multiple skills share a concrete reusable sub-capability, while each still has substantial task-specific logic. Do not use for vague thematic similarity.
- SUBSET: One skill is a strict reusable subskill of another, and the broader skill adds extra workflow, constraints, or context. Do not use for merely related skills in the same domain.
- MERGE: Two or more skills are near-equivalent in triggers, inputs, workflow, and outputs, and should become one canonical skill. Do not use when one skill should delegate to another.

UPWARD_ABSTRACT_INDUCTION: Create an abstract parent skill when peer skills share a higher-level solving pattern or task schema. The parent summarizes composition rather than becoming an executable merged skill.

OUTPUT FOR SCREEN
{
  "stage": "SCREEN",
  "selected": [...],
  "rejected": [...]
}

OUTPUT FOR DECIDE
ASSIGNMENT: <...>
SKILLS: <...>
RELATION_TYPE: <...>
REASON: <...>
END

CONSTRAINTS FOR SCREEN
- selected: candidates worth a relation attempt.
- rejected: candidates not worth it (coarse evidence).
- skills: a pair or small group per entry.
- reason: short, evidence-based.
- No assignment labels, no needs_full_content.

CONSTRAINTS FOR DECIDE
- One DECIDE block per candidate, ready for RELATION BUILDER.
- If coarse evidence is thin, read full skill content first.
- Pick ASSIGNMENT: DOWNWARD (extract) or UPWARD (abstract).
- Fill only the fields for the chosen branch:
    * UPWARD   -> ABSTRACT_PATTERN.
    * DOWNWARD -> RELATION_TYPE, then:
        - SHARED_PART -> SHARED_PART.
        - SUBSET      -> SUBSET_OF (narrower vs broader).
        - MERGE       -> MERGED_SKILL (canonical name).
\end{lstlisting}
\end{verifierbox}

\paragraph{Prompt Template for Relation Builder}
The Relation Builder constructs the assigned relations from full skill contents. For \textit{Downward Atomic Extraction}, it extracts, reuses, merges, or canonicalizes atomic skills and adds explicit reuse references; for \textit{Upward Abstract Induction}, it creates abstract parent skills that capture shared task schemas.
\begin{verifierbox}{Relation Builder Prompt}
\begin{lstlisting}[basicstyle=\ttfamily\small, breaklines=true]
You are the RELATION BUILDER in SKILLPYRAMID. You receive a final assignment from the RELATION ANALYZER and the involved `SKILL.md` contents.

Your job is to construct the actual reusable relation while preserving the skill library's correctness. Be conservative and evidence-grounded.

INPUT

ASSIGNMENT: <DOWNWARD_ATOMIC_EXTRACTION | UPWARD_ABSTRACT_INDUCTION>
SKILLS: <comma-separated skill names>
RELATION_TYPE: <SHARED_PART | SUBSET | MERGE, only for DOWNWARD_ATOMIC_EXTRACTION>
SUBSET_OF: <narrower skill -> broader skill, only for RELATION_TYPE SUBSET>
MERGED_SKILL: <proposed canonical skill name, only for RELATION_TYPE MERGE>
SHARED_PART: <common reusable capability, only for RELATION_TYPE SHARED_PART>
ABSTRACT_PATTERN: <high-level pattern, only for UPWARD_ABSTRACT_INDUCTION>
REASON: <brief evidence-grounded explanation>
END

The caller also provides the full `SKILL.md` contents of the involved skills.

TOP-LEVEL ASSIGNMENTS

DOWNWARD_ATOMIC_EXTRACTION: Build or reuse a lower-level atomic capability. This assignment contains three relation types: SHARED_PART, SUBSET, and MERGE.
- SHARED_PART: Extract the shared reusable operation into one lower-level atomic skill. Then rewrite each source skill so it delegates exactly that shared operation through a reuse reference.
- SUBSET: Rewrite only the broader skill. Wherever it reimplements the covered subskill, replace that local portion with an explicit reuse reference to the narrower skill.
- MERGE: Create one canonical lower-level skill that preserves all useful, non-conflicting details from the near-equivalent source skills. The output must be a standalone skill, not a comparison.

UPWARD_ABSTRACT_INDUCTION: Create one higher-level abstract skill that summarizes the common task schema of the group. Do not merge or replace the children. The abstract skill should describe how lower-level skills compose to solve a class of tasks.

REUSE REFERENCE

Use exactly this format when a skill delegates to another skill:

[reuse skill: <skill name> | when: <short trigger> | provides: <short capability>]

BUILD RULES
- The RELATION ANALYZER assignment is authoritative. Do not change ASSIGNMENT or RELATION_TYPE.
- Do not invent tools, scripts, files, observations, actions, dependencies, or source-skill capabilities.
- Preserve important workflows, constraints, edge cases, and verification checks from source skills.
- For RELATION_TYPE SHARED_PART, the new skill must be narrower than every source skill.
- For RELATION_TYPE SUBSET, rewrite only the broader skill named in SUBSET_OF.
- For RELATION_TYPE MERGE, combine compatible details into one canonical skill and do not preserve duplicate standalone variants.
- For UPWARD_ABSTRACT_INDUCTION, the new skill must be higher-level guidance, not a concrete action script.
- Rewritten skills must remain understandable as standalone `SKILL.md` files.
- Add reuse references only near the affected passages.
- Use UPPERCASE for ASSIGNMENT names, RELATION_TYPE names, and output field labels when labels are required.

OUTPUT FORMAT

Output skill content only. Do not include analysis, markdown fences, JSON, or commentary unless the caller explicitly requests structured metadata.

For generated or rewritten skills, use this format:

name: <kebab-case skill name>
description: <one-sentence description>

# <Skill Title>

## Purpose
<what capability or pattern this skill provides>

## When to Use
<triggers and applicable conditions>

## Inputs / Expected Context
<required observations, task state, objects, files, or prior context>

## Workflow
<step-by-step reusable procedure or high-level composition pattern>

## Verification
<how to confirm progress or completion>

## Constraints / Guardrails
<limits, failure modes, and unsupported assumptions>

## Output
<expected result or capability provided>
\end{lstlisting}
\end{verifierbox}

\paragraph{Prompt Template for Skill Creator}
The Skill Creator generates task-specific skills grounded in the current pyramid. It first uses higher-level skills for \textit{Framework Construction}, then lower-level skills for \textit{Detail Instantiation}; stage-specific retrieval weights implement this priority.
\begin{verifierbox}{Skill Creator Prompt}
\begin{lstlisting}[basicstyle=\ttfamily\small, breaklines=true]
You are the SKILL CREATOR in SKILLPYRAMID. You may be called multiple times.
Each time you are called, you will be clearly informed that you are currently in one of the following two stages.

STAGES
- FRAMEWORK_CONSTRUCTION: Use higher-level or abstract skills first to outline the task structure, subgoals, decision points, success criteria, and high-level solving logic.
- DETAIL_INSTANTIATION: Use lower-level or atomic skills to fill executable operations, required inputs, state checks, verification rules, outputs, and reuse references, then produce the final reusable skill.

Your job is to produce a reusable `SKILL.md` body that is grounded in the supplied task context, retrieved skills, and current environment feedback. Be conservative and evidence-grounded.

INPUT FOR FRAMEWORK_CONSTRUCTION
{
  "stage": "FRAMEWORK_CONSTRUCTION",
  "task_instruction": "<new task instruction>",
  "retrieved_skills": [
    {
      "name": "<skill name>",
      "description": "<skill description>",
      "level": "<optional level or role>",
      "relation_context": "<optional pyramid or graph context>",
      "relevance": "<optional exact, partial, or background relevance summary>"
    }
  ]
}

INPUT FOR DETAIL_INSTANTIATION
{
  "stage": "DETAIL_INSTANTIATION",
  "task_instruction": "<new task instruction>",
  "framework": "<FRAMEWORK_CONSTRUCTION output>",
  "retrieved_skills": [
    {
      "name": "<skill name>",
      "description": "<skill description>",
      "content": "<optional SKILL.md content>",
      "level": "<optional level or role>",
      "relation_context": "<optional pyramid or graph context>",
      "relevance": "<optional exact, partial, or background relevance summary>"
    }
  ]
}

GENERATION PROCESS
- In FRAMEWORK_CONSTRUCTION, use the stage-ranked `retrieved_skills` to extract task class, reusable strategy, decomposition, decision points, and success criteria. Focus on names, descriptions, relation context, relevance summaries, and any supplied high-level procedural content.
- FRAMEWORK_CONSTRUCTION produces a framework only: candidate name, description, task class, high-level logic, subgoals, success criteria, reuse candidates, and cautions. Do not write the final skill body in this stage.
- In DETAIL_INSTANTIATION, use the stage-ranked `retrieved_skills` to extract concrete operations, required inputs, state checks, verification rules, outputs, and reusable failure cautions. Focus on supplied `SKILL.md` content, exact reusable subtasks, and environment-compatible action patterns.
- DETAIL_INSTANTIATION produces the final reusable skill by filling the framework with grounded executable detail and placing reuse references near the substeps they support.
- If retrieved skills are only partially related, reuse their reasoning pattern, verification habits, and cautions, but do not copy unsupported concrete objects, rooms, product IDs, commands, or environment facts.

REUSE REFERENCE
Use exactly this format when a generated skill delegates to another skill:
[reuse skill: <skill name> | when: <short trigger> | provides: <short capability>]

GENERATION RULES
- Follow the current STAGE's output format only.
- Use exact names from retrieved/available skills; invent nothing
  (skills, files, tools, actions, env facts).
- Final skill is self-contained, usable as a SKILL.md body.
- Favor reusable workflow guidance over one-off replay.
- Action verbs/commands only if in the supplied action space.
- Live environment feedback overrides the generated skill.
- If nothing relevant is retrieved, borrow only style and cautions.
- Keep concrete details, constraints, and failure cautions as evidence.
- UPPERCASE for STAGE names, process names, and field labels.

OUTPUT A FOR FRAMEWORK_CONSTRUCTION
STAGE: FRAMEWORK_CONSTRUCTION
NAME: <concise kebab-case candidate skill name>
DESCRIPTION: <one hedged sentence describing when this skill may help>
TASK_CLASS: <reusable task class>
HIGH_LEVEL_LOGIC: <high-level solving logic inferred from abstract or higher-level skills>
SUBGOALS: <ordered subgoals or decision points>
SUCCESS_CRITERIA: <how the final skill should define completion>
REUSE_CANDIDATES: <exact retrieved skill names and the role each may play>
CAUTIONS: <unsupported assumptions, environment dependencies, or failure modes>
END

CONSTRAINTS FOR FRAMEWORK_CONSTRUCTION
- One FRAMEWORK_CONSTRUCTION block.
- No final CONTENT.
- REUSE_CANDIDATES: exact retrieved skill names only.

OUTPUT B FOR DETAIL_INSTANTIATION
Output exactly:
NAME: <concise kebab-case skill name>
DESCRIPTION: <one hedged sentence describing when this skill may help>
CONTENT:
# <Skill Title>
## Purpose
<task class and intended outcome>

## Possible Strategies
<high-level plan, decomposed through abstract skills when relevant>

## Reuse Plan
<reuse references for exact subtask coverage, each near its trigger>

## Execution Checklist
<atomic operations, state checks, and verification habits>

## Cautions
<failure modes, unsupported assumptions, and when to switch tactics>
END

CONSTRAINTS FOR DETAIL_INSTANTIATION
- One final skill, structured by the FRAMEWORK_CONSTRUCTION output.
- Fill operations, inputs, checks, outputs from
  retrieved skills when supplied.
- Reuse references: exact names only, placed near affected passages.
- No analysis, fences, JSON, or commentary outside the skill format.
\end{lstlisting}
\end{verifierbox}

\paragraph{Prompt Template for ALFWorld Evaluation}
The ALFWorld evaluation prompt frames the agent as an embodied household assistant. It specifies the household environment, task goal, valid object-manipulation actions, and the required \texttt{Thought}/\texttt{Action} output format. For skill-based methods, retrieved skills are inserted through the \texttt{\{\{skill\_reference\}\}} field.
\begin{verifierbox}{System Prompt for ALFWorld}
\begin{lstlisting}[basicstyle=\ttfamily\small, breaklines=true]
Interact with a household to solve a task. Imagine you are an intelligent agent in a household environment, and your goal is to perform actions to complete the task.

At the beginning of the interaction, you will be given a detailed description of the current environment and the task goal. At each turn, you will be given the latest observation. You should first think about the current situation and plan your next steps, then output your action for this turn.

Your output must strictly follow this format:

Thought: <your thoughts>
Action: <your next action>

The available actions are:

1. go to {recep}
2. take {obj} from {recep}
3. move {obj} to {recep}
4. open {recep}
5. close {recep}
6. use {obj}
7. clean {obj} with {recep}
8. heat {obj} with {recep}
9. cool {obj} with {recep}

Here, {obj} and {recep} correspond to objects and receptacles in the environment.

After each turn, the environment will provide immediate feedback. Use this feedback to plan your next steps. If the environment outputs "Nothing happened", the previous action was invalid, and you should try another valid option.

{{skill_reference}}

Your response should use the following format:

Thought: <your thoughts>
Action: <your next action>
\end{lstlisting}
\end{verifierbox}

\paragraph{Prompt Template for ScienceWorld Evaluation}
The ScienceWorld evaluation prompt frames the agent as an assistant conducting scientific experiments. It defines the available rooms, experiment-oriented actions, object inspection and manipulation operations, and the required \texttt{Thought}/\texttt{Action} output format. Retrieved skill references are inserted through the \texttt{\{\{skill\_reference\}\}} field when applicable.
\begin{verifierbox}{System Prompt for ScienceWorld}
\begin{lstlisting}[basicstyle=\ttfamily\small, breaklines=true]
You are a helpful assistant conducting scientific experiments in an environment.

The environment contains several rooms: kitchen, foundry, workshop, bathroom, outside, living room, bedroom, greenhouse, art studio, and hallway.

You should explore the environment and find the items needed to complete the experiment.

You can teleport to any room in one step.

All containers in the environment have already been opened, so you can directly get items from the containers.

The available actions are:

open OBJ: open a container
close OBJ: close a container
activate OBJ: activate a device
deactivate OBJ: deactivate a device
connect OBJ to OBJ: connect electrical components
disconnect OBJ: disconnect electrical components
use OBJ [on OBJ]: use a device or item
look around: describe the current room
examine OBJ: describe an object in detail
look at OBJ: describe a container's contents
read OBJ: read a note or book
move OBJ to OBJ: move an object to a container
pick up OBJ: move an object to the inventory
pour OBJ into OBJ: pour a liquid into a container
mix OBJ: chemically mix a container
teleport to LOC: teleport to a specific room
focus on OBJ: signal intent on a task object
wait: take no action for 10 steps
wait1: take no action for one step

{{skill_reference}}

Your response should use the following format:

Thought: <your thoughts>
Action: <your next action>
\end{lstlisting}
\end{verifierbox}

\paragraph{Prompt Template for WebShop Evaluation}
The WebShop evaluation prompt frames the agent as a web-shopping assistant. It provides the current observation and available actions at each step, restricts actions to \texttt{search[keywords]} and \texttt{click[value]}, and requires the \texttt{Thought}/\texttt{Action} output format. Retrieved skill references are inserted through the \texttt{\{\{skill\_reference\}\}} field for skill-based methods.
\begin{verifierbox}{System Prompt for WebShop}
\begin{lstlisting}[basicstyle=\ttfamily\small, breaklines=true]
You are web shopping.

I will give you instructions about what to do. You must follow the instructions.

At each round, I will give you an observation and a list of available actions. You should choose one valid action based on the current state and the instruction.

You may use the search action if search is available. You may also click one of the available clickable buttons.

An action must follow one of these formats:

search[keywords]
click[value]

The search keywords are up to you, but the value in click[value] must exactly match one of the available actions.

If the action is invalid, nothing will happen.

Design search keywords carefully so that they are specific enough to find the requested product.

{{skill_reference}}

Your response should use the following format:

Thought: <your thoughts>
Action: <your next action>
\end{lstlisting}
\end{verifierbox}

\end{document}